\documentclass[nonacm, authorversion]{acmart}
\AtBeginDocument{%
  \providecommand\BibTeX{{%
    \normalfont B\kern-0.5em{\scshape i\kern-0.25em b}\kern-0.8em\TeX}}}

\usepackage{tcolorbox}
\newtcolorbox{mybox}{colback=red!5!white,colframe=red!75!black}
\usepackage{arydshln}

\usepackage{dsfont}

\begin{document}

\title[The Long Arc of Fairness]{The Long Arc of Fairness: Formalisations and Ethical Discourse}

\author{Pola Schw{\"o}bel}
\authornote{Both authors contributed equally to this research.}
\email{posc@dtu.com}
\affiliation{%
  \institution{Technical University of Denmark}
  \streetaddress{Anker Engelunds Vej 1}
  \city{Copenhagen}
  \country{Denmark}}

\author{Peter Remmers}
\authornotemark[1] 
\affiliation{%
  \institution{Technische Universität Berlin}
  \streetaddress{Straße des 17. Juni 135}
  \city{Berlin}
  \country{Germany}}
\email{peter.remmers@tu-berlin.de}


\begin{abstract}
In recent years, the idea of formalising and modelling fairness for algorithmic decision making (ADM) has advanced to a point of sophisticated specialisation. However, the relations between technical (formalised) and ethical discourse on fairness are not always clear and productive. Arguing for an alternative perspective, we review existing fairness metrics and discuss some common issues. For instance, the fairness of procedures and distributions is often formalised and discussed statically, disregarding both structural preconditions of the status quo and downstream effects of a given intervention. We then introduce \emph{dynamic fairness modelling}, a more comprehensive approach that realigns formal fairness metrics with arguments from the ethical discourse. A dynamic fairness model incorporates (1) ethical goals, (2) formal metrics to quantify decision procedures and outcomes and (3) mid-term or long-term downstream effects. By contextualising these elements of fairness-related processes, dynamic fairness modelling explicates formerly latent ethical aspects and thereby provides a helpful tool to navigate trade-offs between different fairness interventions. To illustrate the framework, we discuss an example application -- the current European efforts to increase the number of women on company boards, e~.g.~ via quota solutions -- and present early technical work that fits within our framework.

\end{abstract}




\maketitle

\section{Introduction} \label{sec:intro}

At the core of fair machine learning (fair ML) research lies the question:\textit{What is fairness?} A fundamental goal of research in fair ML is to define ethical standards for ML technologies and to help build tools that live up to such standards. This endeavor has become urgent in light of the rapid advancements within ML which have enabled the widespread use of algorithmic decision making (ADM). The stakes are high, both for society and for individuals, and some of these decision making systems have failed in dramatic and systematic ways: racially biased ADM in the criminal justice system \cite{angwin2016machine}, facial recognition failing on women of color \cite{buolamwini2018gender} , sexist hiring \cite{dastin2018} and racist search engines \cite{noble2018algorithms} are only some notorious examples.

Especially important for auditing the fairness of algorithms are fairness metrics, i.e. formal criteria by which to score fairness, and a multitude of metrics has been proposed. However, despite great research efforts and a plethora of approaches, there are fundamental issues that the field has not, so far, been able to overcome. As noted by Jacobs et al. \cite{jacobs2021measurement} and Binns \cite{binns2020apparent}, some of these issues are consequences of the tendency of fair ML research to conflate formal analysis of fairness with the discussion of ethical principles. While certain basic ideas of fairness are formally constructed as fairness metrics, these formalisms are then analysed too narrowly without entering a (non-formal) \textit{ethical} debate. For example, formal contradictions between two different fairness metrics have been construed as technical flaws of the metrics, when in fact both fairness metrics are perfectly valid formalisations of ethical principles (as is the case for the apparent conflict between individual and group fairness \cite{binns2020apparent}, see further discussion in Sec. \ref{sec:existing_metrics}).

In the line of reasoning of this contribution, we argue for a clarification of the roles of formal contributions and ethical debate in fair ML research. Rooted in quantitative fields, fair ML depends on formalisations. However, analysis of formalised criteria alone cannot determine the grounds for a choice between different criteria (other than their formal properties, e.g.~ whether they are consistent with each other, or whether they have computational properties such as differentiability which make them suitable as a loss function). Fair ML thus also depends on a comprehensive discussion of ethical principles, goals and values. We aim to develop a formalisation strategy that incorporates such ethical considerations, and show that such formalisations can aid the non-formal debates in turn. 




By reviewing existing fairness metrics and their weaknesses in Section \ref{sec:existing_metrics}, we identify a second issue in the current fairness debate: Constructed fairness criteria are often not sufficiently contextualised. Procedural fairness criteria assume that an unbiased decision process alone will lead to a fair state of the world (see Sec. \ref{sec:procedural_criteria}). Operating under this assumption, they neither acknowledge nor adjust for biased data and as a consequence are prone to reinforcing existing inequalities. Outcome based fairness modelling provides a more promising approach, however, the standard outcome based metrics usually fail to capture ethically relevant structural differences between groups (see Sec. \ref{sec:outcome_criteria}). That is, they do not investigate \textit{how} a certain outcome distribution arose, or the long-term effects of intervening in such a distribution. 

In contrast, we argue that we should optimistically demand of fair ML exactly this: It should be thought of as a tool to \textit{intervene} in the status quo and improve conditions for the previously disadvantaged. 
The transformation of established bureaucratic procedures towards automation-based processes offers historically unique opportunities for reevaluation and restructuring society. Such transformation holds the promise of improving structural conditions for historically disadvantaged groups and individuals via access to, for example, better jobs, wealth and education. 
In this context, a \emph{fairness intervention} is a procedure that is specifically designed to address and intervene into pre-existing discrimination. A well-known general example for this strategy is the practice of positive (or `affirmative') action. In contrast to the paradigm of `blind' decision-making that intentionally excludes certain protected features from the process, respective decisions explicitly take demographic differences into account in order to counteract historic forces of discrimination. We call this motive for fair ML \textit{interventional}.


With this context we will introduce a framework we call \textit{dynamical fairness modelling} in Section ~\ref{sec:dyn_modelling}. Dynamical fairness modelling, we argue, helps bridging the gap between (un-)fairness in the world (the ethical discourse) and formalisations (the formal discourse) via the following mechanism. It forces the decision maker to explicate their long term \textit{goals} in ethical terms (as opposed to the merely implicit ethical dimension of a predominantly technical choice), their \textit{formalisation} as well as the expected long-term \textit{effects} of any suggested interventions. Rather than evaluating fairness interventions statically and in isolation, dynamic fairness modelling prompts the decision maker to reflect on, model and evaluate the downstream effects of any chosen decision policy -- according to the interventional perspective. 
To illustrate the framework, we will discuss a conceptual as well as a computational example. After reviewing some existing technical work in the direction we propose, we will conclude by returning to a more philosophical treatment in Section \ref{sec:conclusion} where we will also discuss limitations.


\section{Procedural and outcome-based fairness metrics} \label{sec:existing_metrics}
In this section, we briefly present and discuss some technical approaches that are prominent in the debates on fair ML, although we do not claim to give an exhaustive overview.  The approaches can be categorised as procedural and outcome-based criteria of fairness, roughly following the distinction between pure procedural justice and perfect procedural justice as introduced by Rawls (\cite{rawls1999theory} p. 74-75). We will demonstrate how discussions of `static' formal metrics lead to issues that can be addressed by incorporating further context that is initially not present in the existing formalised criteria. We will eventually arrive at a contextualised modelling approach, dynamical fairness modelling, in Section \ref{sec:dyn_modelling}. 


\subsection{Procedural Criteria}
\label{sec:procedural_criteria}



Procedural fairness is determined by criteria that refer to the \emph{process} of a decision (as opposed to the \emph{outcome} of a decision). Procedural fairness criteria may follow the ethical principle to treat everyone equally in a decision process, independently of any specific attributes.\footnote{According to the ethical goal of `equal treatment', sufficiently random decisions may be considered fair insofar as probabilities are equal for everyone (cf. \cite{creel2021algorithmic} for an interesting discussion of the (un-)fairness of random decisions.).} On the other hand, decision making procedures are \emph{unfair} if they follow principles that are themselves ethically unacceptable, independently of the outcome. Specifically, considering given histories and structures of discrimination, it may be ethically unacceptable to base a decision on certain sensitive attributes, for example using attributes like race or gender in the context of hiring. The so-called `blindness' approach to anti-discrimination as formalised in the `fairness through unawareness' criterion constructs a decision procedure that is supposed to be fair by simply not considering any such protected attributes \cite{gajane2017formalizing}.\footnote{Protected attributes are features such as religious affiliation, age or sexual orientation that are `protected' by law of many countries. Individuals cannot be discriminated against based on such attributes for example in the context of hiring (e.g.~ US Civil Rights Act Title VII) or housing  (e.g.~ US Fair Housing Act).} For example, the principle of `color blindness' refers to racial categories: `Generally, color blindness minimises the use and significance of racial group membership and suggests that race should not and does not matter.' (\cite{plaut2018color}, p. 200). 

\paragraph{Why procedural criteria fail: They neither acknowledge nor adjust for biased data.}

There are fundamental problems with this approach to anti-discrimination. Although the procedural constraints of the `blindness' approach might be effective in preventing direct (i.e. explicit) discrimination, other variables can act as proxies for protected attributes. In this case, there is information flowing from the protected attribute $A$ to the outcome even if the category of $A$ is not explicitly used by the model. This happens because structural discrimination is statistically effective in many ways: It correlates protected attributes like race or gender to geographic residence, socioeconomic status, education, medical records, family background, criminal records and other attributes. Consequently, if there is discrimination, its effects are very likely manifest in data. And if the data that correlates to protected attributes is used in a decision making procedure, the process may be as discriminating as if the protected attributes were explicitly used in the first place.

This problem is exacerbated in ML-based ADM. Machine learning works by extracting patterns from large amounts of historical data by statistical inference, referred to as `training the model', and then using these patterns to determine decisions. Considering this fundamental mechanism by which ML works, it becomes clear that ML can never be better than the data used to train it. We can at best hope that the algorithm perfectly captures the information we have presented it with. But if we train an algorithm on biased training data, it will reflect such biases. 

For this reason, the principle of non-discrimination as `exclusion of protected attributes' is a formal criterion that does neither acknowledge nor adjust biased data. At best, a decision procedure realises equal chances and opportunities for everyone affected. At worst, a procedure mirrors data bias and proliferates discrimination. In this case, a `blind' decision procedure reproduces a given distribution of capabilities and opportunities that was unfair in the first place. 

\begin{table}
\centering
\begin{tabular}{|p{4.5cm}|p{4.5cm}|p{5cm}| } 
 \hline
 \textbf{Fairness principle} & \textbf{Fairness metric (name)} & \textbf{Definition} \\ \hline \hline
\textbf{Fair process} \textit{(Rawls' pure procedural justice \cite{rawls1999theory})} & & \\ \hdashline
  `Blindness': Protected attribute should not be used in the decision. & Fairness through Unawareness &   Protected attributes are not explicitly used in the prediction process \cite{gajane2017formalizing},  $F(X, A) = F(X)$.\\ \hline
Protected attribute should not cause the decision. & Counterfactual Fairness \cite{kusner2017counterfactual} &  $p(F | do(A = 0)) = p(F | do(A = 1))$, $do(\cdot)$ is the do-operator which denotes an intervention on the protected attribute. \\  \hline
& (Un-)Resolved Discrimination, Proxy Discrimination \cite{kilbertus2017avoiding}  & Causal paths between ethically relevant variables and outcome are (un-)blocked, see \cite{kilbertus2017avoiding}. \\ \hline \hline
\textbf{Fair outcome } \textit{(Rawls' perfect procedural justice \cite{rawls1999theory})} & & \\ \hdashline
 No subjective discrimination: Qualified people
should be equally likely to obtain the job/mortgage/etc. across groups. & (Formal) Equal Opportunity \cite{hardt2016equality} & $p(F | A = 0, Y = 1) =  p(F |A = 1, Y = 1)$  \\ \hline
 In addition: \textit{Un}qualified people should also be equally likely to \textit{not} get the job/mortgage/etc. across groups. & Equalised Odds \cite{hardt2016equality} & $p(F  | A = 0, Y = y) = p(F | A = 1, Y = y)$ for $\ y \in \{0, 1\} $  \\ \hline
 Equal representation, diversity & Demographic Parity & $p(F | A = 0) = p(F | A = 1)$ \\ \hline
 `Treat like cases alike' \textit{(Aristotle)} & Individual Fairness & $D(F(x_1), F(x_2)) \leq d(x_1, x_2)$  for $D$ and $d$ distance functions in the output and input space. \\ \hline 
\end{tabular}
\caption{Some fairness principles and their formalisations; for the relationship between fairness principles and ethical goals see Sec. \ref{sec:dyn_modelling}. In the right column the notation is as follows. $F$: the predictor (with slight abuse of notation, this can refer to both a single function as well as a distribution of outcomes, i.e.\ we do not properly distinguish here between deterministic and probabilistic algorithms), $X$: the (distribution of) features, $A$: a protected attribute (e.g.\ gender or race), $Y$: the (distribution of) true labels (e.g.\ whether someone is qualified for the job/mortgage/etc.).} \label{table:goals_formalisations}
\end{table}


\paragraph{Possible solutions}
To deal with the issue of proxies, we might try to somehow `filter' the data used in the process in a more elaborate way. An example for this approach works by identifying which data (other than the data that explicitly refers to the protected attributes) should or should not be used by a decision making algorithm. For example, Grgić-Hlača et al. \citep{grgic2018beyond} propose an approach based on surveying users about whether any feature should be used in a fair decision making process. On the other hand, not all features that correlate with the protected attribute might be unacceptable for ethical decision-making: For example, a job might require the applicant to hold an engineering degree; and holding such a degree is positively correlated with being male in many countries. As a consequence, less women and non-binary individuals\footnote{Like much of the existing fair ML literature that we build on, we acknowledge the use of overly simplistic categories and false binaries in this work. We view efforts towards inclusive categories and intersectionality as absolutely necessary, and as an orthogonal research direction to static vs. dynamical fairness modelling which is the focus here.} might be hired without the gender attribute being used in a discriminatory way. In other words, only causal relations between protected attributes and decision outcomes are problematic in terms of fairness, and only those need to be corrected for. This idea is explored in causal fairness approaches (e.g. \cite{kilbertus2017avoiding, kusner2017counterfactual, chiappa2018causal}). While such approaches provide elegant solutions where causal data is available (i.e. where we know the reason why a certain situation came about), there is a reason why modelling is usually done in the observational sense, based on correlation rather than causation: It is in most cases difficult, if not unfeasible, to model the full causal process leading to a certain feature distribution; e.g. the cultural and historic reasons for women and non-binary individuals not to choose undergraduate degrees in engineering in the first place remain unexamined.

Instead, we can acknowledge present and historic discrimination that result in biased data and work towards ways to address them. A fairness intervention should be thought of as a procedure that is specifically designed to address and intervene into pre-existing discrimination. Corresponding to this \textit{interventional perspective}, we argue for \textit{dynamical fairness modelling} which we will introduce in Sec.~\ref{sec:dyn_modelling}.


\subsection{Outcome-based criteria} \label{sec:outcome_criteria}
The above considerations and further examples from the literature on bias and fairness suggest that we should eventually judge the fairness of the procedure by its outcome. While a certain procedure may seem completely unbiased and non-discriminating by itself, it may appear differently when we look at its outcomes \cite{trautmann2016process}. Perhaps we find out that although a seemingly fair decision procedure carefully precludes sensitive data, it still leads to an apparently unfair distribution of opportunities and goods. Consequently, decision procedures that incorporate potentially biased data should be evaluated by looking at the outcomes. Outcomes can be measured in terms of the distribution of goods, e.g.~ resources and material goods, but also opportunities, capabilities and well-being. Fairness then correlates to the ethical acceptability of a certain outcome. The ethical goal of a respective fairness intervention could be an equal distribution of goods or, alternatively, a distribution that is proportional to a certain merit. In this setting, an algorithm’s fairness can be evaluated by reference to the distribution of outcomes it produces, i.e.~ the state of a world in which decisions were made according to the algorithm’s predictions or recommendations. Generally, outcome-based approaches are suited to bypass the previously mentioned blind spot of procedural fairness, because an evaluation of outcomes is based on criteria of fairness that are to some extent detached from the bias of the original data. For this reason, these approaches seem to be motivated by the idea of controlling potential unfairness by actively neutralizing certain biases (although the interventional stance will turn out to be a more adequate point of view).

\paragraph{Group fairness metrics}
Early contributions to algorithmic fairness propose outcome-based criteria such as demographic parity \cite{calders2009building} or equality of opportunity \cite{hardt2016equality} (see Table \ref{table:goals_formalisations} for formalisations of these and other metrics). Applied to the example of hiring and gender, demographic parity corresponds to hiring the same proportion of male and female candidates. Equality of opportunity requires hiring at the same proportions conditioned on the candidates' qualifications. In our example, \textit{qualified} male and female applicants should be hired at the same rates.
The difference between demographic parity and equality of opportunity becomes apparent when considering the case of unequal qualification rates between the genders. If indeed more qualified men apply, the latter criteria allows for differences in hiring rates, where the first does not. Because the protected attribute is usually thought to encode the membership to a demographic group (gender, race, etc.), criteria based on such attributes are summarised under the term group fairness.

\paragraph{Individual fairness metrics}
Seemingly in contrast to group fairness are so-called individual fairness metrics. According to individual fairness, a decision is fair if similar individuals are treated the same way, or, in terms of Aristotle’s account of justice, that similar cases are treated alike. In our example of hiring, to satisfy individual fairness, equally qualified candidates should either both be hired or not hired, regardless of which group they are categorised in. Much effort in technical work on fair ML focuses on evaluating different fairness metrics against each other, and proving various incompatibility statements \cite{chouldechova2017fair, dwork2012fairness}. Formally, apart from very specific cases, group fairness and individual fairness can not be satisfied simultaneously. If the underlying distribution of features is different between demographic groups, we cannot obtain demographic parity while at the same time treating individuals from both groups the same. In order to achieve demographic parity, we need to allow for preferential treatment of the less qualified group. Binns \citep{binns2020apparent} resolves this conflict by pointing to the shared underlying ethical goal of both individual and group fairness; we briefly discuss his work below.

\paragraph{Why outcome-based criteria fail: They do not acknowledge structural differences between groups}

 Individual fairness requires a measure for similarity; mathematically speaking, we need a metric to define the distance of individuals $x$ and $y$ in the input space (assume that $x$ belongs to protected group $X$ and $y$ to $Y$, respectively). In the hiring example, the metric would be defined in terms of some qualification score, and would typically ignore protected attributes when determining similarity. Proceeding in this way, one implicitly decides that belonging to group $X$ or $Y$ is ethically irrelevant for the decision at hand, following the principle of `blindness' as described in Sec. ~\ref{sec:procedural_criteria}. But from the interventional perspective, this stipulation is misleading, because we are interested \textit{specifically} in socio-economic, historic and structural differences between groups. Instead of merely ignoring unwanted data that correlates to protected attributes as in the `blindness' approach, individual fairness should rather construct relevant similarities between selected attributes. A good similarity metric should reflect ethically relevant differences.\footnote{For example, in a Rawlsian luck-egalitarian sense, a decision should correct for circumstances negatively affecting an individual’s qualification score that lie outside their control.} 
 Interestingly, as Binns \cite{binns2020apparent} shows, when using a similarity metric that accounts for ethically relevant differences between groups, individual and group fairness can become commensurable. Designing such a more holistic similarity metric is not trivial as any choice is necessarily rooted in ethical reasoning and underlying values. Indeed, we need to explicate our ethical stance: `conflicts are not primarily the result of selecting individual or group fairness measures. Instead, they are likely to be the result of unstated conflicting moral and empirical assumptions regarding the decision-making context' (\cite{binns2020apparent}, p. 519). 


Metrics like equality of opportunity or equalised odds suffer from a similar shortcoming: They do not account for the different realities of protected groups. The two metrics define unfairness as an unfair distribution of errors, i.e.~ when opportunities are wrongfully denied for people of certain demographic groups. However, as Eidelson \cite{eidelson2021patterned} argues, perfectly accurate, i.e. error-free, decisions can be unfair as well if they occur in the context of what he terms \textit{patterned inequality} between groups. As an example, imagine a bank giving out loans. A lending decision is considered accurate whenever the lender can repay. Being wealthy should make it easier to pay back the loan; if the investment does not go as planned, there might be alternative income streams to alleviate the loss and pay back the bank. Thus, an algorithm which only approves loans to wealthy people will be highly accurate, as individuals from this group will likely be qualified in the sense of being able to repay. However, by employing such a decision criterion, people born into less wealthy families will never be afforded the opportunity of taking out a loan to make an investment, say, in their own business, in order to improve the economic situation for themselves. The effect is especially dire in cases where different socio-economic factors are linked (e.g. wealth and race) such that entire communities are systematically excluded from opportunity. Note that this is not a problem specific to machine learning or automation in general, but of merit-based decision making overall. As Kasy and Abebe \cite{kasy2021fairness} state: `under this perspective, inequality [...] is acceptable if it is justified by merit [...], no matter where the inequality [in merit] is coming from'.

Demographic parity seems specifically designed to break such patterned inequalities. It may require drastic positive action, for example approving bank loans at equal rates for men and women. But this can have negative consequences for individuals that belong to the very group that is supposed to benefit, because it ignores the unfortunate reality of the gender pay gap, women's lower wages on average, and thus their potentially lower ability for paying back a loan. Receiving a bank loan that one is unable to repay, however, leads to less financial well-being, a worse credit score and eventually being worse off than without having received the loan in the first place. This, of course, is not to say that the consequences of requiring demographic parity are always negative. More often than not it will be hugely beneficial for an individual to be afforded an opportunity. Nevertheless, the potential harms of group fairness metrics like demographic parity or equalised odds for those groups that are supposed to benefit should be reflected in the implementation of ADM. 

To summarise this section, we identify two general approaches to fairness: procedural and outcome-based approaches.
Procedural criteria fail to account for existing biases in data and are therefore prone to reproducing existing inequalities. Within the category of outcome-based approaches, we discuss different fairness metrics, formally divided into group fairness metrics and individual fairness metrics. Group fairness metrics entail certain risks for the groups that are supposed to benefit from them. As purely distributive criteria, group fairness metrics neither address nor explicitly control the conditions that lead to a certain distribution. For example, a distribution according to demographic parity is not in itself valuable, but only if it helps to change the social conditions that contribute to the development of strongly disparate distributions in the first place. That means that we should not only discuss fairness in terms of (static) distributions between different groups, but as a result of \textit{processes} that shape and determine these distributions. 
\paragraph{Strategy}
We conclude that a process-based, i.e.~ dynamical modelling perspective is necessary to meaningfully reason about fairness in the interventional sense -- a perspective that many existing metrics are lacking. We have also seen that existing approaches often suffer from a lack of explicitly stated ethical goals. The necessary ethical debate is sometimes conflated with and obscured by the formal debate, such as in the discourse of the apparent conflict between group and individual fairness metrics. As a consequence, we formulate the following desiderata for fairness modelling:

\begin{enumerate}
    \item The ethical goals should be stated explicitly, and independently of formalisation. 
    \item Any intervention should be evaluated based on its impact towards ethical goals, i.e.~ whether it improves the conditions underlying disparate distributions of goods between demographic groups. 
\end{enumerate}

The following Section \ref{sec:dyn_modelling} will develop a framework for fairness modelling according to these considerations. 

\section{Dynamical fairness modelling} \label{sec:dyn_modelling}
We will now outline the implementation of the dynamical fairness modelling framework, first in a short overview (Sec.~ \ref{sec:implementation}), and then with an example. Observing a mainly US-centric debate, we will work with a European case study: gender quotas on company boards as a potential measure to reduce gender inequality in the workforce. This measure has been discussed and/or implemented in multiple European countries such as Norway,\footnote{Norway is not a itself a EU member state, but has re-kindled the positive action debate across the European Union when it introduced a minimum requirement of $40$\% of women on all company boards of publicly listed companies as early as $2006$.} Belgium, Italy, France, Germany and the Netherlands \cite{ekin2018}; California followed in September 2018 (CA Senate Bill 826, \cite{gertsberg2021gender}). After this conceptual example (in Sec.~\ref{sec:example}), we will illustrate what a computational implementation of the framework can look like. To do so, we will review existing technical approaches for dynamical fairness modelling, in particular the pivotal 2019 work by Liu et al.\ \cite{liu2018delayed}  (Sec. ~\ref{sec:computational_example} and \ref{sec:existing_work}). 

\subsection{Implementing Dynamic Fairness Modelling} \label{sec:implementation}

\paragraph{(1) Explicate Ethical Goals}
The first element of our proposed modelling framework is an explication and discussion of the long-term goals in ethical terms, i.e. independent of possible formalisations. While these explications will likely refer to existing philosophical principles of fairness or justice, e.g. to positions like egalitarianism or equality of opportunity, what we call `ethical goals' is meant to be more concrete and contextualized, especially w.r.t.~ the long term effects of any possible intervention. A discussion on the level of abstract ethical principles or ideals is often insufficient because fairness interventions are rarely just applications of general rules to specific problems. 
Instead, the explication of a specific ethical goal should refer to a given background of structural discrimination and inequality, ideally by incorporating the specific histories and conditions that are relevant for the context of the projected decision making system. To complement general principles, localized knowledge about racism, sexism, colonialism or classism etc. should play a role in the discussion of ethical goals. Additionally, these reflections should be very specific in terms of those local contexts that will be influenced and transformed by the development of an ADM system.

We give two brief examples for ethical goals here that will be elaborated in the rest of this section. 
Firstly, consider the realization of the value of diversity in the assembly of teams. A concrete manifestation of `diversity' will depend on which groups were previously un- or underrepresented and why. For example, when reasoning about women in the workplace, it is useful to consider factors such as the traditionally higher workload for women in the home (see case study in Sec. \ref{sec:example}). Another example is an institution setting the goal to actively advance substantive equality of opportunity  between demographic groups. A good fairness intervention might aim to help those that are structurally disadvantaged due to the local history and culture, not only by affording them opportunities directly but by helping them to successfully compete for those (see example in Sec. ~\ref{sec:computational_example}). 


\paragraph{(2) Formalisation} In a second step, decision makers approximate a formalisation of the previously explicated ethical goals. In a simple case, this formalisation might simply correspond to one of the existing fairness metrics. 
For example, as Binns \cite{binns2020apparent} shows convincingly, an egalitarian ethical stance could be formalised in terms of both group or individual fairness metrics. (Formal) equality of opportunity corresponds to the fairness metric of the same name. Ethical goals around diversity and equal representation can mathematically be expressed via the demographic parity metric. The ethical principle to `treat like cases alike' which is in many contexts required by legislation can be encoded via the individual fairness metric \cite{dwork2012fairness}. Table \ref{table:goals_formalisations} contains some fairness principles and their formalisations; they are discussed in more detail in Sec.~ \ref{sec:existing_metrics}.

Applying existing fairness metrics in this sense is an easy way to arrive at formalisations of ethical goals; however,  they should not always be expected to correlate to existing metrics as easily. As argued in the previous paragraph (1), our ethical goals usually require a higher level of specificity and contextualisation. In particular, as we will show with an example in the next section, many ethical goals are more robustly formalised under a \textit{long-term} view. This temporal perspective is important to address not only the symptoms of structural discrimination, but also the conditions that produce them. This dimension is not expressed in the standard fairness metrics, which is why we call them `static'. Under the dynamic modelling point of view, additional formalisations become available. For example, as we will see in Sec.~ \ref{sec:existing_work}, Liu et al.\ \citep{liu2018delayed} suggest optimising for equally distributed features (rather than outcomes) as a proxy for fairness. 


\paragraph{(3) Modelling Down-Stream Effects}  Once the decision maker has formalised their ethical goal, they can start to evaluate any potential course of action against it. This means developing a mathematical model of the downstream consequences of a given action, e.g.\ will admitting more female students to university programs increase the number of qualified female applicants for certain positions. Of course, the quality of this model is essential for the success of our approach, i.e. it should be based on empirical research and expert knowledge of the problem at hand. Early technical work on dynamical modelling of algorithmic fairness usually proposes models based purely on assumptions which is also valuable, at least to investigate the framework. \\

Having broken down the dynamical modelling pipeline, we note that a main advantage lies in its explicitness and, consequently, transparency. Each of the steps corresponds to stating or formalising assumptions in a way that can readily be critiqued and tested. Critiquing the first step corresponds to asking:  Do we agree with this ethical goal? Disagreeing about the notion of fairness or justice corresponds to a philosophical debate with multiple stakeholders, and ethicists being domain experts. The second step can be evaluated by asking: Does our formalisation indeed capture the ethical goal we have stated?  As \cite{jacobs2021measurement} points out, such measurement modelling tasks are standard problems in the quantitative social sciences. They can be accomplished by, for example, testing whether the formalisation is consistent in the sense of test–retest reliability: If the same `fairness-test' comes out differently for very similar scenarios, the operationalisation at question is not robust, a sign of a poor measurement. For the last step we ask: Does a given fairness intervention indeed have the claimed effect? Again, this question can, in principle, be answered with expert knowledge and empirical data whenever the research is available. For example, Kalev et al. \cite{kalev2006best}, survey the effect of a variety of positive action policies on management diversity. If such data is not available yet one might decide to roll out the intervention and measure its effects (given budget and ethical constraints). By enabling us to challenge underlying assumptions and mechanisms dynamical modelling provides an interface for interdisciplinary collaboration between stakeholders, technologists, ethicists, social scientists and other experts.

\subsection{An Example: Women on Company Boards} \label{sec:example}
The EU as well as individual member countries have been concerned with gender inequality in the workplace and have discussed and implemented a range of interventions, most notably gender quotas for company boards. Such quotas require the boards of publicly listed companies in the respective countries to contain at least a certain percentage of women, typically between $30$ and $40$\% where such solutions are implemented  \cite{jourova2016gender}. This section illustrates dynamical fairness modelling by measures of such a fairness intervention. 
We note that hiring decisions for company boards are not algorithmic in the sense of being fully automated or processed by machines -- certainly, such high stakes personnel decisions are currently made by humans. Rather, they are algorithmic in a broader sense that there is `a step-by-step procedure for solving a problem or accomplishing some end'\footnote{Definition of algorithm according to \url{https://www.merriam-webster.com/dictionary/algorithm}.}, i.e.\ an underlying set of (implicit) rules that the decision makers are following. In this sense, most `principled' decisions can be considered algorithmic. 

\paragraph{(1) Ethical Goals: Equality of Opportunity, Diversity and Representation}
The 2013 report on `Positive Action Measures to Ensure Full Equality in Practice between Men and Women, including on Company Boards' \cite{selanec2013positive} prepared for the European commission identifies three ethical goals (referred to as `normative goals' in the text) of such interventions. The first goal is to  `improve the ability of the disadvantaged group to compete for the available opportunities', i.e.\ ensuring substantive equality of opportunity. \textit{Substantive} (or, in Rawls' terms, \textit{fair}) equality of opportunity is distinct from \textit{formal} equality of opportunity, in that it does not require equal hiring criteria on paper, but equality in the \textit{chances to satisfy} those criteria \cite{sep-equal-opportunity}. Secondly, they aim `to limit the negative effects on women’s position in the labour market of the unequal distribution of responsibilities in the family'. The third goal is to `to ensure the balanced representation of men and women in bodies with significant decision-making powers'. Instead of taking the individual's perspective, this last goal is formulated from society's point of view. It could be interpreted as the value of diversity in itself, via some sort of democratic legitimacy (i.e.\ bodies of significant decision-making powers should be demographically representative of the people they are governing) or via the improved results achieved by diverse teams \citep{page2019diversity}.

\paragraph{(2) Formalisation: Demographic Parity, but in the Long Term}
At a first glance, the third ethical goal seems to translate into a formalisation straightforwardly: `balanced representation of men and women' corresponds to demographic parity. That is, if the base population consists of $50$\% women, one would aim for the same proportion of female board members. 
When comparing with the notion of demographic parity encoded in EU legislation, we note a subtle difference to the classic notion from the fair ML literature where demographic parity is understood to apply to \textit{any single decision in isolation}. However, in this real world example it is usually formulated as a long-term goal, i.e.\ quotas are to be met within a time frame, typically a small number of years \citep{jourova2016gender}. While this might seem like a political technicality at first (we cannot force companies to fire and hire new boards on the spot), we believe that we see a general property of fairness principles and their formalisations at play: 
They are often best thought of as aspirational long-term goals rather than short-term strategies. Indeed, in our example, people generally agree that more balanced company boards are desirable in the long-term, but disagree on the best measures to achieve such parity: In $2010$ \citep{selanec2013positive}, `$77$\% of the Europeans are of the opinion that we need more women in management positions [...]. At the same time, Europeans are rather sceptical about strong positive action measures. The Eurobarometer survey found that $44$\% of European respondents ($44$\% W, $44$\% M) consider that the most efficient measures consist of encouraging enterprises and public administrations to take measures to foster equality between women and men (``code of good practice") and to fight against stereotypes’ while ‘concerning the imposition of quotas by law, it is favoured by $19$\% of European respondents ($20$\% W, $18$\% M)'. 

\paragraph{(3) Modelling Down-Stream Effects: The Effectiveness of Positive Action}  
Having decided on demographic parity (formalised in the long-term sense) as the ethical goal, states can consider different policies to achieve them. A naive way might be to immediately require demographic parity, i.e.\ re-appoint company boards in a gender-balanced manner and keep the demographic parity constraint for all future personnel decisions. While fulfilling the criterion on paper, this approach does not seem to actually align with many people's moral intuitions (as seen in the Eurobarometer survey data above). They might disagree with this intervention for the reasons that we are familiar with from the fair ML literature: Assume the reason for seeing few female board members is not blatant sexism, or what economists call taste-based discrimination \cite{becker1957economics}, but rather the applicant pool containing few qualified women according to the current hiring criteria. Then, achieving demographic parity immediately can imply hiring `less qualified' women, or, given equal qualification, preferring women to men which might seem unfair towards their `more qualified' male counterparts. Such violation of the equal treatment principle is discussed in the fair ML literature as a contradiction between individual and group fairness (see  Sec.~\ref{sec:outcome_criteria}). We note that contrary to this intuition, EU law explicitly allows for preferential treatment in the context of positive action: `With a view to ensuring full equality in practice between men and women in working life, the principle of equal treatment shall not prevent any Member State from maintaining or adopting measures providing for specific advantages in order to make it easier for the underrepresented sex to pursue a vocational activity or to prevent or compensate for disadvantages in professional careers' (Article 157(4), Consolidated version of the Treaty on the Functioning of the European Union (TFEU)).

Secondly, as we have seen in Sec. ~ \ref{sec:outcome_criteria}, some argue that drastic preferential treatment might have negative consequences for the women themselves: Women might be perceived as less competent in their jobs when quotas are employed in their selection regardless of their actual qualifications \cite{dematteo1996evaluations}. If they were indeed appointed despite being less qualified, they might be less likely to being re-appointed or recommended by their colleagues for other opportunities. More dramatically for the underrepresented group, under-qualified women in such jobs might lead to statistical discrimination against the group of women as a whole. After observing less qualified female individuals, decision makers might conclude that women in general are less able to perform well in the job. We note that this argument is based on the implicit assumption that there are essentially no qualified females in the candidate pool (since we would still be able to hire the most qualified ones under a quota solution). This seems implausible given the fact that more women than men graduate from higher education programs in the EU: In 2019, $46$ \% of women aged $30-34$ had attained tertiary education and only $35$ \% of men across the EU Eurostat (2021). \footnote{Gender statistics. Eurostat. Retrieved from \url{https://ec.europa.eu/eurostat/statistics-explained/index.php?title=Gender_statistics\#Education} } On the other hand, the `negative example' argument works in the other direction as well: quotas and the resulting increased representation of women produce more role models and can lead to an increased willingness for women to compete for the jobs \citep{balafoutas2012affirmative}.

Thirdly, it is not clear that demographic parity is indeed desirable if it is achieved by continuously applying quotas. Fairness interventions are lawful and desirable, but they should tackle the \textit{cause} of the inequalities and should thus be temporary. According to the UN Convention on Elimination of Discrimination against Women (Article 4(1)), positive action measures `shall be discontinued when the objectives of equality of opportunity and treatment have been achieved' \cite{freeman2012convention}. The goal of a good fairness intervention is that it will become redundant over time. 

Interestingly, operationalising demographic parity on boards in this sense illustrates a problem with the ethical goal and its formalisation itself. If we achieve demographic parity by continuing to apply quotas every time we have to make a hiring decision but the actual distribution of qualified candidates never changes, i.e.\ the parity never becomes the `natural' state (or the stationary distribution of the process), the strategy does not actually appear to be successful in achieving equality in the workplace. Instead, we want to improve the situation for the underrepresented group and design interventions which actually lead to more women being qualified for those board seats (this might not mean changing the women but changing the qualifications). Thus, the ethical goal in its first formulation above, to `improve the ability of the disadvantaged group to compete for the available opportunities', turns out to be a more complete picture. Once this is achieved, we can obtain demographic parity without any further interventions because more women will be qualified \footnote{This is based on the assumption that women and men have similar cognitive markups and would, given `free' choice choose similar professions in distribution. This is a somewhat controversial assumption (what if women choose not to be on board seats?), but it seems to be consistent with the EU law's conception of gender equality.}. The formalised debate has, in a way, informed the ethical one (rather than just vice versa). 

Under these considerations, one might define `robust', long-term demographic parity as goal and develop other, temporary strategies to achieve it by improving women's conditions for competing in the labour market. The range of such alternative strategies is wide: A group of approaches aims to enable mothers to (re-)join the workforce, those include flexible work hours or part-time employment, or providing company childcare facilities. Some countries require nomination parity, i.e. employers have to nominate two candidates, one of each gender for every position \cite{selanec2013positive}. We also might invest more in developing female talent early on, in universities or graduate programs, or invest in diversity training or more inclusive job ads.
Amongst those interventions, we naturally prefer those which are most effective, i.e.\ the best under our model of down-stream effects. In this example, the first group of measures seems to be the least effective \cite{selanec2013positive}. Wheatley, 2016 \cite{wheatley2017employee} suggests that part- and flexi-time arrangements often have negative effects on women's careers since they are likely to re-enforce the traditional dynamics of women working more in household and families. \\

This example has illustrated the strengths of the dynamical modelling perspective: As argued theoretically in Sec. ~ \ref{sec:existing_metrics}, we have seen how ethical deployment of any algorithmic decision making system in a complex, real world required its embedding into a much bigger context than what static fairness metrics can provide. If we aim to implement it as a fairness intervention, i.e.\ under the interventional perspective from Sec.~\ref{sec:intro}, we need to consider any decision's consequences over time, and how those feed back into the features relevant for decisions in the future. The dynamical fairness modelling approach can be a formal language for this. Indeed, it can be helpful for such reasoning by bridging the gap between the ethical and the formal debate. 
After illustrating the approach conceptually, we will now move on to the technical perspective of how dynamical fairness modelling might be implemented in a computational setting by reviewing early existing work. 




\subsection{A Computational Example: Liu at al.'s Delayed Impact of Fair Machine Learning} \label{sec:computational_example}




Liu et al. \cite{liu2018delayed} propose a mechanism which allows for temporal analysis of ML decision processes by introducing `a one-step feedback model of decision-making that exposes how decisions change the underlying population over time'. Under this model, the authors study whether certain fairness criteria indeed improve the of well-being of a disadvantaged group, or whether they might even lead to a decline in a variable of interest.
To our knowledge, this is some of the earliest technical work that fits within the framework of dynamical fairness modelling, and we will illustrate here how Liu et al.'s approach is one strategy to implementing it. 

\textit{Ethical goal:} Their ethical goal is to `promote the long-term well-being of disadvantaged groups' (\cite{liu2018delayed}, p.~1). 

\textit{Formalisation:}
Two groups $A$ and $B$ associated with a protected attribute are characterised by distributions $\pi_{A/B}$ of qualification scores $\mathcal{X}$. The notion of well-being referred to in the ethical goal is then equated with this qualification score. For example, the authors use an individual's credit score as a proxy for their financial well-being in the lending example. Institutions have selection policies  $\tau_{A/B}$ (rates at which score they accept credit applications), and those have down-stream effects on the individuals. In particular, they assume the availability of a function $\Delta : \mathcal{X} \mapsto \mathds{R}$ that provides the expected change in score for a selected individual at a given score. The expected change for the group as a whole is denoted by $\Delta \mu_{A/B}$. The authors then distinguish between long-term improvement ($\Delta \mu_{A/B} > 0$), stagnation ($\Delta \mu_{A/B} = 0$), and decline ($\Delta \mu_{A/B}< 0$) for the groups $A$ and $B$. The suggested metric for evaluating a decision making policy refers to the change of this average qualification score. A desirable policy leads to an increased average qualification score for the individuals of the disadvantaged group. 

\textit{Model of downstream effects}: 
The authors assume access to a function $\Delta: \mathcal{X} \mapsto  \mathds{R}$ that provides the expected change in score for an individual with score $x$. As discussed in Sec.~ \ref{sec:dyn_modelling}, such a function is in practice difficult to construct. In their lending example, they assume the following simple structure: They denote by $\rho(x)$ the probability of an individual with score $x$ to be able to repay the loan. $c_+$ is the benefit from being granted a loan and being able to repay, $c_-$ is the cost for the individual of defaulting on the loan (for example, the worsened credit score). Then, $ \Delta(x) = c_+ \rho(x) + c_- (1 - \rho(x))$.

The authors show that static fairness metrics, especially demographic parity, can under certain conditions lead to a decline of the qualification score, and thus to the protected group being worse off in the long term. This finding ties in with the problems outlined in Sec. ~\ref{sec:existing_metrics}. As a corrective, the authors suggest optimising for an improvement of the qualification score for the disadvantage group directly rather than applying existing fairness metrics after the fact. This suggestion perfectly aligns with our framework. Instead of deciding on interventions beforehand and evaluating their consequences, we suggest to work backwards from the goal. 

\subsection{Related Work} \label{sec:existing_work}
Similar in spirit to Liu et al. \cite{liu2018delayed}, Zhang et al.~ \citep{zhang2020fair} discuss the impact of static fairness metrics and constraints on the long term well-being of different demographic groups. Unlike Liu et al.'s work, an individual’s qualification is here modelled as a latent, unobservable variable. Observable scores like school grades are viewed as noisy estimates for the underlying qualification -- a relevant difference in the light of ongoing debates about discriminating bias of school grades or standardised tests \cite{eberle1989sat, geiser2020sat}. Their findings again highlight the complexity of the issue: Whether a given, static fairness constraint is beneficial or detrimental downstream depends on the specifics of the problem at hand and cannot be determined without an analysis of the decision’s consequences over time.

Kannan et al.~\cite{kannan2019downstream} analyse fairness policies for college admission and share our view on the essence of the fair ML issue: `What is often unstated (and perhaps not even explicitly considered by the colleges) is what exactly the long term goals of these policies are, beyond the short term goal of having a diverse freshman class' (\cite{kannan2019downstream}, p. 2). The authors formalise two such long term goals by analyzing how the college admission and grading policy influences a potential employer’s hiring decision. Firstly, downstream equal opportunity requires that suited college graduates are equally likely to be hired independent of their demographic. Secondly, elimination of downstream bias demands that `rational employers selecting employees from the college population should not make hiring decisions based on group membership' (\cite{kannan2019downstream}, p. 2). This second criterion is equivalent to demanding that the college grades are distributed such that the employer can apply the `blindness' criterion from see Sec.~\ref{sec:outcome_criteria} without obtaining sub-optimal decisions, i.e. hiring candidates with subpar qualifications. Like Zhang et al.~\citep{zhang2020fair}, this work models a student’s true qualification as a latent variable that can only be estimated noisily by standardised test scores or college grades. Their finding consists in yet another ‘inconsistency statement’: in general, downstream equal opportunity and elimination of downstream bias cannot be achieved simultaneously.

Heidari et al.~ \cite{heidari2021allocating} take a societal perspective rather than focusing on the individual. The authors formalise a mathematical model for allocating opportunities such as college admissions to people. Motivated by an extremely strong correlation between US parents’ and their kids’ socioeconomic status (thus low intergenerational mobility), the effects of positive action on intergenerational socioeconomic status is analysed. In line with our intuition about dynamic fairness modelling and the importance of a long-term view, the authors find the following: An optimal allocation policy that only takes the current generation into account will not employ positive action. However, when future generations are taken into account the optimal policy -- in the utilitarian sense of maximising the number of people who are given an opportunity and succeed -- will include positive action. Intuitively, this is because a child of a well-off individual is likely to be well-off themselves, and so giving somebody the chance to improve their socioeconomic status has positive downstream effects for society.


\section{Conclusion} \label{sec:conclusion}


This work has introduced a framework for dynamical fairness modelling, which we believe to have two main advantages over many of the existing fairness metrics. Firstly, it forces the decision maker to \textit{explicate their ethical goals} and commitments, hereby increasing transparency and helping to disentangle the formal and the ethical debates underlying fair ML. This clarification is motivated by the observation that most problems of fairness cannot be solved in the context of a purely technical discussion. While formalisation and technical implementation of fairness metrics may clarify important aspects, the results remain too limited to address ethical and political issues. Thus, we want to foster a technical debate which is rooted in, and informed by, an ethical one. 
Secondly, it provides a more \textit{contextualised} approach than existing methods. In particular, it accounts for biased data (as a consequence of inequalities in the status quo) and it provides a better starting point for addressing structural differences between groups, eventually improving the conditions for the previously disadvantaged. We have identified this motive for fair ML as the interventional perspective. 

In our thinking about technology's role in the process, we perceive an opportunity. This opportunity, we believe, is not aimed at technological `solutionism': While a technological approach cannot count as a 'solution' by itself, it can work to suggest a certain level of discourse -- specifically, a translation of technical metrics into terms compatible to an ethical assessment (and vice versa). We have seen this interplay of different levels of discourse in Sec. ~\ref{sec:example}, where modelling efforts have aided our ethical reasoning. Thus, we propose dynamical fairness modelling as a technically mediated way to present issues of fairness in more appropriate terms.\footnote{On the mediating role of technology, see \cite{verbeek2011moralizing}.} 

\paragraph{Limitations} The core of our framework is a model of the downstream effects of any fairness intervention. Developing such a model is difficult. How does a college admission, bank loan or hiring decision today affect an individual’s well-being, qualifications and socio-economic status in the future? One might argue that if we had access to such information, we might already be much better at designing fair policies. In the context of ML, this information could come in the form of datasets recording populations over time. Such datasets are not currently part of the standard machine learning toolbox, but could easily be made available given the `big data' culture and ways we collect large amounts of data on essentially everything. Of course, the issues around privacy and the economy of surveillance practices arising from this type of data collection themselves pose a set of ethical questions. 

We note that our modelling approach is relevant to a certain type of decision maker. A somewhat broad scope is required for taking the interventional perspective, both in terms of goals/motivations and competencies. Dynamical fairness modelling is relevant to decision processes that happen on a relatively long timeline, and are aiming to make societal change. Decision makers in a public institution or the government come to mind, and we deem the framework equally relevant from a research perspective. On the other hand, it might be less applicable for actors within companies which structurally are often operating on shorter time horizons, and whose primary goals might be different from changing society. But for those striving to make real change in the `interventional sense' of improving conditions for the previously disadvantaged, we hope this contribution is useful. 

\newpage

\bibliographystyle{ACM-Reference-Format}
\bibliography{sample-base}

\end{document}